\documentclass[journal]{vgtc}

\onlineid{0}

\vgtccategory{Research}
\vgtcpapertype{System, Analytics & Designs}

\shortauthortitle{Park \MakeLowercase{\textit{et al.}}: NeuroCartography}

\author{
Haekyu Park,
Nilaksh Das,
Rahul Duggal,
Austin P. Wright,
Omar Shaikh,
Fred Hohman,
Duen Horng (Polo) Chau
}

\authorfooter{
    The authors are at Georgia Institute of Technology.
    E-mail:\\
    \{haekyu, nilakshdas, rahulduggal, apwright, oshaikh, fredhohman, polo\} @gatech.edu.
    Fred Hohman is currently at Apple.
}

\ifpdf
  \pdfoutput=1\relax
  \usepackage{graphicx}
  \DeclareGraphicsExtensions{.pdf,.png,.jpg,.jpeg}
\else
  \usepackage{graphicx}
  \DeclareGraphicsExtensions{.eps}
\fi%

\usepackage{microtype}
\PassOptionsToPackage{warn}{textcomp}
\usepackage{textcomp}
\usepackage{mathptmx}
\usepackage{times}

\usepackage{cite}
\usepackage{tabu}
\usepackage{booktabs}
\usepackage{xcolor}
\usepackage{mathtools}
\usepackage{amsfonts}
\usepackage{enumitem}
\usepackage{xspace}
\usepackage[normalem]{ulem}

\newcommand{\inception}[0]{InceptionV1}
\newcommand{\embeddingview}[0]{Neuron Projection View\xspace}
\newcommand{\sideview}[0]{Neuron Neighbor View\xspace}
\newcommand{\graphview}[0]{Graph View\xspace}
\newcommand{\clusterpopup}[0]{Cluster Popup\xspace}
\newcommand{\cascade}[0]{Concept Cascade\xspace}
\newcommand{\presim}[0]{SimTopImgs}
\newcommand{\actsim}[0]{SimActMap}

\newtheorem{definition}{\textit{Definition}}

\definecolor{red}{RGB}{198,50,42}
\definecolor{agreen}{RGB}{74, 198, 148}
\definecolor{purple}{RGB}{158, 62, 177}
\definecolor{aqua}{RGB}{87, 180, 181}
\definecolor{orange}{RGB}{255,143,40}
\definecolor{amber}{rgb}{1.0, 0.75, 0.0}
\definecolor{awesome}{rgb}{1.0, 0.13, 0.32}
\definecolor{bronze}{rgb}{0.8, 0.5, 0.2}
\definecolor{indigo}{rgb}{0.0, 0.25, 0.42}
\definecolor{heliotrope}{rgb}{0.87, 0.45, 1.0}
\definecolor{forestgreen}{rgb}{0.13, 0.55, 0.13}
\definecolor{ginger}{rgb}{0.69, 0.4, 0.0}
\definecolor{jade}{rgb}{0.0, 0.66, 0.42}
\definecolor{mediumslateblue}{rgb}{0.48, 0.41, 0.93}
\definecolor{mint}{rgb}{0.24, 0.71, 0.54}
\definecolor{mulberry}{rgb}{0.77, 0.29, 0.55}
\definecolor{linkColor}{RGB}{6,125,233}

\newcommand{\todo}[1]{\textcolor{red}{[TODO: #1]}}
\newcommand{\hide}[1]{}
\newcommand{\haekyu}[1]{\textcolor{purple}{[#1 -Haekyu]}}
\newcommand{\nilaksh}[1]{\textcolor{agreen}{[#1 -Nilaksh]}}
\newcommand{\rahul}[1]{\textcolor{mulberry}{[#1 -Rahul]}}
\newcommand{\austin}[1]{\textcolor{jade}{[#1 -Austin]}}
\newcommand{\omar}[1]{\textcolor{awesome}{[#1 -Omar]}}
\newcommand{\fred}[1]{\textcolor{aqua}{[#1 -Fred]}}
\newcommand{\polo}[1]{\textcolor{orange}{[#1 -Polo]}}

\newcommand{\add}[1]{\textcolor{blue}{#1}}
\newcommand{\remove}[1]{{\textcolor{red}{\sout{#1}\,}\normalfont}}

\newcommand{\mkclean}{
  \renewcommand{\todo}[1]{}
  \renewcommand{\haekyu}[1]{}
  \renewcommand{\nilaksh}[1]{}
  \renewcommand{\rahul}[1]{}
  \renewcommand{\austin}[1]{}
  \renewcommand{\omar}[1]{}
  \renewcommand{\fred}[1]{}
  \renewcommand{\polo}[1]{}
  
  \renewcommand{\add}[1]{{##1\normalfont}}
  \renewcommand{\remove}[1]{}
}

\mkclean

\newcommand{\model}{\textsc{NeuroCartography}\xspace}

\title{NeuroCartography: Scalable Automatic Visual Summarization of Concepts in Deep Neural Networks}

\vgtcpapertype{System, Analytics & Designs}

\abstract{
Existing research on making sense of deep neural networks often focuses on neuron-level interpretation, which may not adequately capture the bigger picture of how concepts are collectively encoded by multiple neurons.
We present \model{}, an interactive system that scalably summarizes and visualizes concepts learned by neural networks. 
It automatically discovers and groups neurons that detect the same concepts, and describes how such neuron groups interact to form higher-level concepts and the subsequent predictions.
\model{} introduces two scalable summarization techniques:
(1) \textit{neuron clustering} groups neurons based on the semantic similarity of the concepts detected by neurons (e.g., neurons detecting ``dog faces'' of different breeds are grouped); and
(2) \textit{neuron embedding} encodes the associations between related concepts based on how often they co-occur (e.g., neurons detecting ``dog face'' and ``dog tail'' are placed closer in the embedding space).
Key to our scalable techniques is the ability to efficiently compute all neuron pairs' relationships, in time linear to the number of neurons instead of quadratic time.
\model{} scales to large data, such as the ImageNet dataset with 1.2M images.
The system's tightly coordinated views integrate the scalable techniques to visualize the concepts and their relationships, projecting the concept associations to a 2D space in \embeddingview{}, and summarizing neuron clusters and their relationships in \graphview{}.
Through a large-scale human evaluation, we demonstrate that our technique discovers neuron groups that represent coherent, human-meaningful concepts.
And through usage scenarios, we describe how our approaches enable interesting and surprising discoveries, such as concept cascades of related and isolated concepts.
The \model{} visualization runs in modern browsers and is open-sourced.
}


\teaser{
  \centering
  \includegraphics[width=0.95\linewidth]{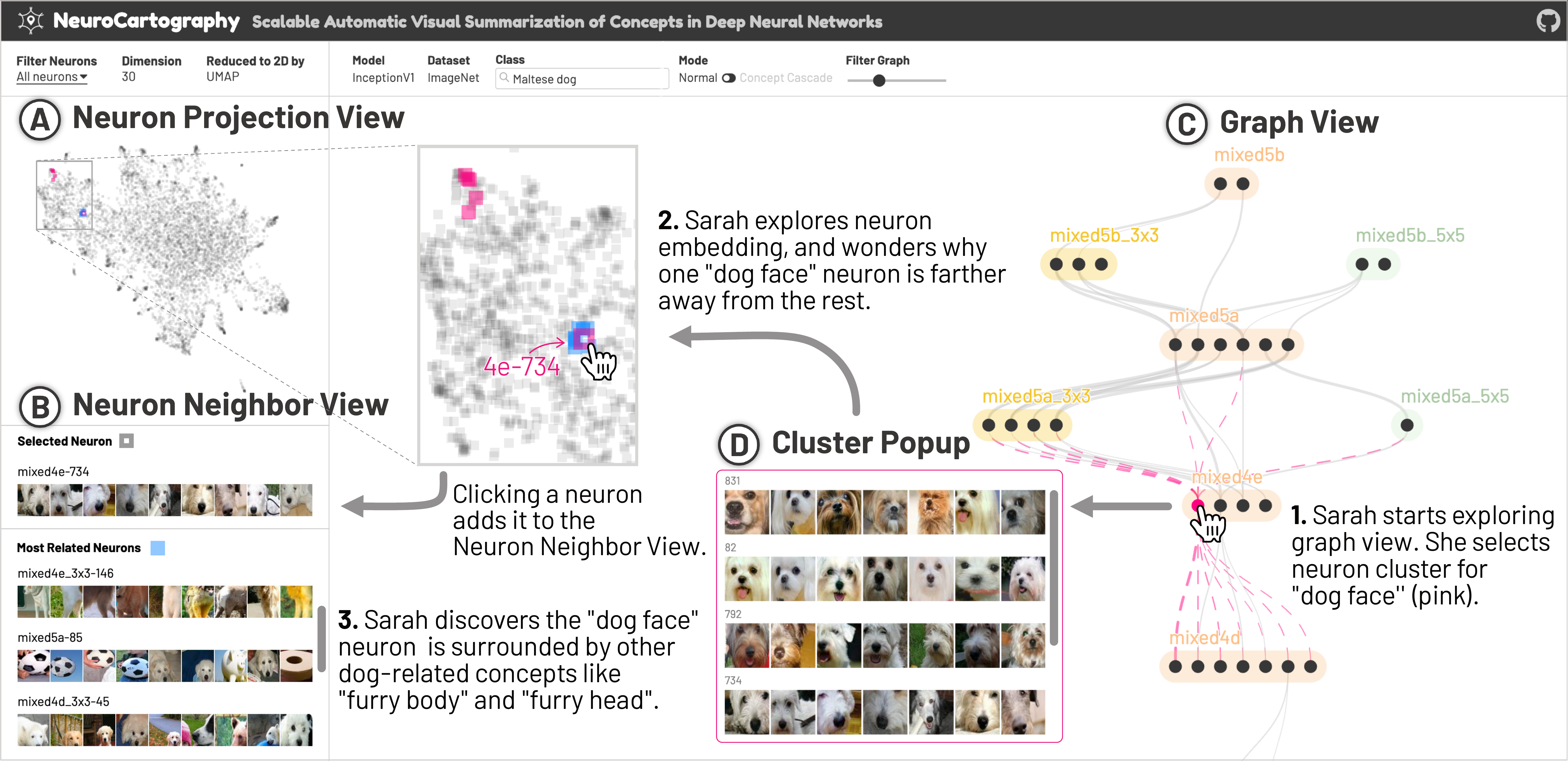}
  \vspace{-3mm}
  \caption{
    \model{} scalably summarizes concepts learned by deep neural networks, by automatically discovering and visualizing groups of neurons that detect the same concepts.
    \textbf{1.} Our user Sarah starts exploring which neuron clusters (shown as circles) play an important role for \inception{} to predict the \textit{Maltese dog} class, through the \textit{\graphview{}} (at C), which clusters neurons based on the semantic similarity of the concepts detected by the neurons, and shows how those concepts interact to form higher-level concepts.
    Selecting the neuron cluster for ``dog face'' (in pink) visualizes its member neurons' features with example patches in the \textit{\clusterpopup{}} (at D).
    \textbf{2.} Member neurons are highlighted in the 
    global \textit{\embeddingview{}} (at A), which summarizes all neurons' concepts from all layers by projecting them on a 2D space, placing related concepts closer together;
    Sarah wonders why one ``dog face'' neuron (\#734) is farther away from the rest.
    \textbf{3.} Adding that neuron to the \textit{\sideview{}} (at B) enables discovery of the most related neurons 
    (nearby blue neurons in projection), such as ``furry body'' (neuron 4e\_3x3-146) and ``furry head'' (4d\_3x3-45), suggesting that the proximate projection region is
    in fact capturing dog-related concepts. Concepts of neurons and neuron groups are manually labeled.
    \label{fig:teaser}
    \vspace{-3mm}
    }
}

\vgtcinsertpkg

\begin{document}

\firstsection{Introduction}
\maketitle
\begin{figure}[t]
    \centering
    \includegraphics[width=0.75\columnwidth]{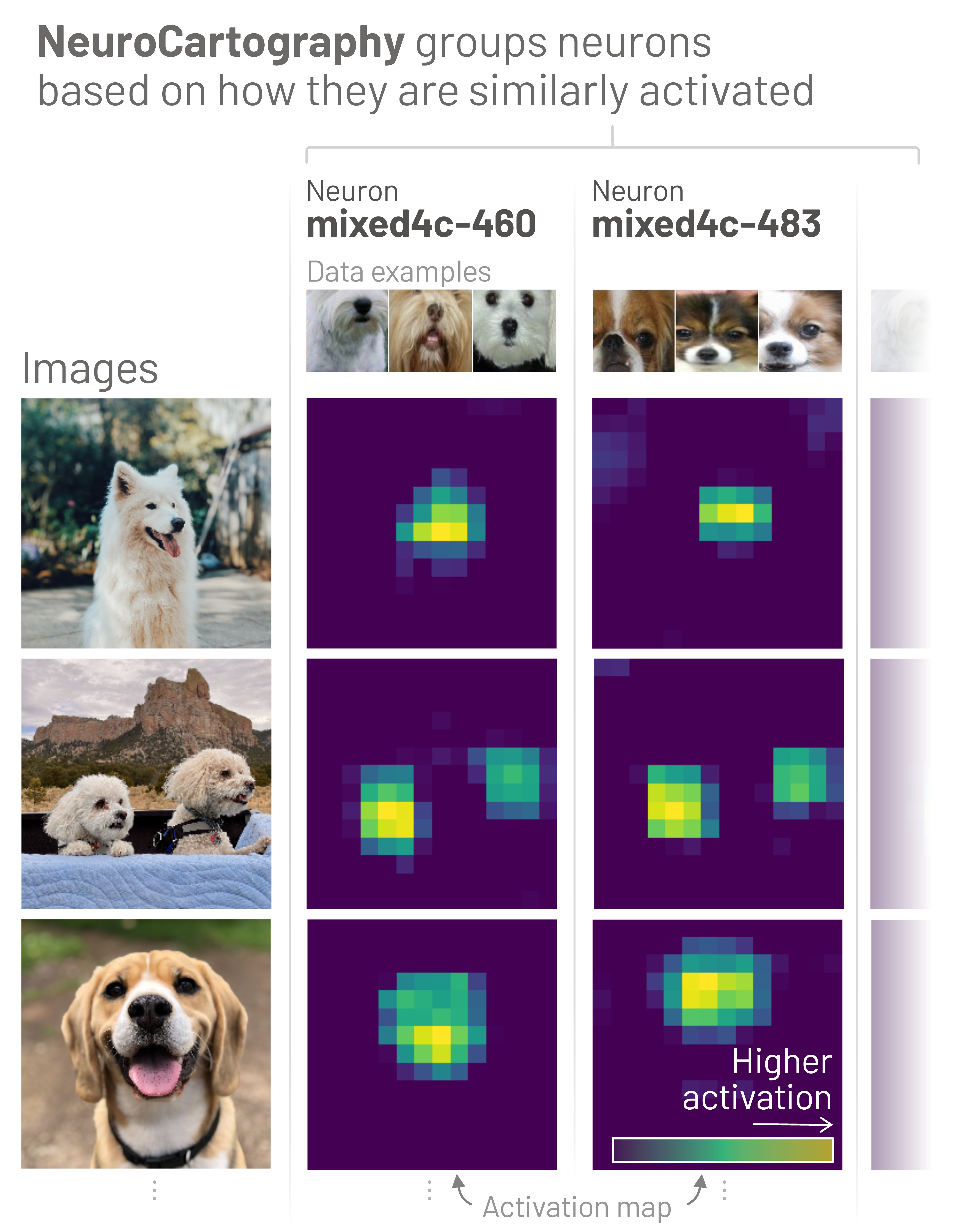}
    \vspace{-3mm}
    \caption{
        To summarize the concepts learned by a DNN,
        \model{} groups neurons based on how 
        similarly they are activated, e.g., by the ``dog face'' concept.
        Here, neurons 460 and 483 in layer mixed4c of \inception{} model are similarly activated by the ``dog face'' concept, and are grouped in the same cluster by our approach.
    }
    \vspace{-6mm}
    \label{fig:neuron-selectivity}
\end{figure}

Deep Neural Networks (DNNs) have demonstrated remarkable success in many applications, 
such as object detection~\cite{zhao2019object, liu2020deep}, 
speech recognition~\cite{deng2013new, noda2015audio}, 
and data-driven health care~\cite{ravi2016deep, rajkomar2018scalable}. 
However, they are often considered \textit{opaque} due to their complex structure and large number of parameters. 
To help practitioners and researchers more confidently and responsibly deploy machine learning models,
there have been major efforts and calls from both government and industry to enable greater model interpretability ~\cite{wright2020comparative, hohman2020summit}.
There is research that aims to explain models' predictions 
based on the inputs, such as regions of input images that contribute the most to the models' predictions~\cite{ghorbani2019towards, kim2018interpretability, szegedy2016rethinking, selvaraju2017grad, simonyan2013deep}.
However, such techniques often do not describe \textit{how} and \textit{where} the input features are used  within the model. 
Recent research posits a key step towards answering these questions is to interpret \textit{neurons} (also called \textit{channels}), since they are highly activated for specific features from the input data~\cite{bau2020understanding, olah2017feature,hohman2019s,das2020bluff, das2020massif}.
While \textit{neuron-level} interpretation may be a promising approach to discover insights, inspecting individual neurons can be time-consuming; furthermore, individual inspection does not easily reveal how \textit{multiple} neurons may detect the same features, which means users could easily miss the bigger picture of the DNNs' decision-making processes.
For example, \autoref{fig:neuron-selectivity} shows that the ``dog face'' concept is detected by multiple neurons in \inception{} model.
Although it is a well-documented phenomenon that
multiple neurons detect similar features~\cite{olah2020overview} (especially in model pruning research~\cite{he2017channel, jaderberg2014speeding, wen2016learning, duggal2020rest}),
there is a lack of research in 
(1) developing scalable summarization techniques to discover concepts collectively learned by multiple neurons,
and (2) enabling users to interactively interpret such concepts and their similarities. 
\model{} aims to fill this critical research gap. 

\smallskip
\noindent
\textbf{Contributions.} In this work, we contribute:
\begin{itemize}[itemsep=0mm, topsep=0mm,parsep=1mm, leftmargin=3mm]
    \item 
        \textbf{\model{}, an interactive system that scalably summarizes and visualizes fundamental concepts} that contribute to the behaviors of large-scale image classifier models (\autoref{fig:teaser}), such as \inception{}~\cite{szegedy2015going}.
        \model{} automatically discovers groups of neurons that detect the same concepts and describes how such neuron groups interact to form higher-level concepts and the subsequent predictions (\autoref{sec:interface}).
    \item 
        \textbf{Two scalable concept summarization techniques}:
        (1) \textit{neuron clustering} groups neurons based on the semantic similarity of the concepts that neurons detect (e.g., neurons detecting ``dog faces'' of different breeds are grouped); and
        (2) \textit{neuron embedding} encodes the associations between related concepts based on how often they co-occur
        (e.g., neurons detecting ``dog face'' and ``furry body'' are placed closer in the embedding space, as seen in \autoref{fig:teaser}A, B).
        Both efficient techniques avoid naively comparing all neuron pairs, resulting in a time complexity that is linear to the number of neurons, rather than quadratic time.
        Our techniques scale to large data, such as ImageNet ILSVRC 2012 with 1.2M images~\cite{russakovsky2015imagenet} (\autoref{sec:method}).
    \item 
        \textbf{Interactive exploration of \cascade{}} 
        enables users to selectively initialize and examine how a concept detected by a neuron group would trigger higher-level concepts across subsequent layers in a neural network.
        \model{} visualizes the user-selected concept's ``cascading effect,'' 
        helping users interpret the 
        successive concept initiations and relationships (\autoref{sec:cascade}).
    \item
        \textbf{Empirical findings through large-scale human evaluation and discovery scenarios.}
        Through a large-scale human evaluation,
        we demonstrate that \model{} detects neuron groups representing coherent concepts with consistent meaningful human interpretations (\autoref{sec:study}).
        We describe how \model{} can help discover several interesting and surprising ﬁndings through usage scenarios, like identifying concept cascades for related classes (e.g., dogs of different breeds) and identifying isolated concepts that are unrelated to all other concepts in a neural network (\autoref{sec:scenario}).
    \item 
        \textbf{An open-sourced, web-based implementation} that helps broaden people’s access to neural network interpretability research without the need for advanced computational resources. 
        Our code and data are open-sourced\footnote{\textcolor{linkColor}{\url{https://github.com/poloclub/neuro-cartography}}}, and the system is available at the following public demo link:
        \textcolor{linkColor}{\url{https://poloclub.github.io/neuro-cartography/}}.
\end{itemize}

\section{Related Work: Neural Network Interpretability}

A Deep Neural Network (DNN) takes as input a data instance and outputs its class label.
Transforming the input to the output may potentially involve billions of numerical computations, the scale of which is incomprehensible to humans.
Recent research aims to summarize these calculations into abstract concepts, helping humans interpret how DNNs work.
In this section, we provide an overview of existing research for neural network interpretability that motivate our work.

\subsection{Interpretation at Input-, Layer-, and Neuron-Level}

\label{sec:neuronlevel}
A common approach to interpret how DNNs operate internally is to study \textbf{features} detected by the models.
Saliency Maps~\cite{Simonyan14a} and Grad-CAM~\cite{selvaraju2017grad} explain a DNN's prediction at the \textit{input-level}, by finding contributing features from the input images. 
TCAV~\cite{kim2018interpretability} explains a feature's importance for a class at the \textit{layer-level}, by vectorizing activations in a layer and measuring the vector's sensitivity to the classes.
A growing number of techniques~\cite{bau2020understanding, hohman2019s, fong2018net2vec, nguyen2016multifaceted} aim to interpret such features at the \textit{neuron-level}.
Given a predefined segmented image that highlights regions for human-interpretable concepts, Net2vec~\cite{fong2018net2vec} represents these concepts as the linear combination of thresholded activation maps from activated neurons.
These techniques quantitatively determine how each neuron contributes to the concept image.
Feature-visualization techniques~\cite{olah2017feature, nguyen2016multifaceted, nguyen2016synthesizing}, on the other hand, algorithmically generate synthetic images that maximally activate a particular neuron, thus visualizing what features are preferred by each neuron.
Other methods~\cite{olah2017feature} also generate examples cropped from real images in the dataset that highly activate a specific neuron.
In \model{}, we use dataset examples to represent features detected by each neuron. 

Some research focuses on interpreting the \textbf{relationships among features} in the form of embedding vectors \cite{karpathytsne, carter2019activation, nguyen2016multifaceted}.
Karpathy's t-SNE feature representation~\cite{karpathytsne} visualizes similarity among features at the \textit{input-level}, by projecting each image onto 2D space using the image's DNN features (from a full-connected layer).
Nguyen et al.~\cite{nguyen2016multifaceted} represent multifaceted features of a neuron at the \textit{input-level}.
For example, it embeds onto 2D space all bell pepper images that activate a neuron detecting bell peppers,
revealing the neuron's facets (e.g., red, green, yellow peppers).
Activation Atlas~\cite{carter2019activation} computes embeddings at the \textit{layer-level},
where each embedded point represents a linear combination of all neurons' activation strengths in a layer, based on spatial patches extracted from those neurons' activation maps.
As the above techniques generate embeddings at abstraction levels different from ours, it is unclear how they may directly map each point in the embedding space back to specific neurons. In other words, existing approaches cannot pinpoint which neurons are responsible for detecting specific features, which \model{} supports (\autoref{sec:embedding}).

\subsection{Semantic Similarity of Neurons}
Recent research~\cite{olah2020overview, he2017channel, duggal2019cup, jaderberg2014speeding, wen2016learning} suggests that neurons tend to detect similar features.
\autoref{fig:neuron-selectivity} illustrates this claim by highlighting how different neurons detect the same feature of a dog's nose.
Olah et al.~\cite{olah2020overview} highlights many examples of similar neurons in InceptionV1 and visualizes which concepts are detected by such neurons; however, the examples are manually curated by the authors.
Identifying neurons that discover similar concepts also has practical benefits: in the neural network compression community, several methods~\cite{duggal2019cup, he2017channel, jaderberg2014speeding, wen2016learning, duggal2020rest} leverage potential neuron redundancies to generate compressed models while maintaining prediction accuracy.
Even though these methods can measure neurons' similarity, there is limited work in interpreting their \textit{semantic} similarity.
CNNPruner~\cite{li2020cnnpruner}, an interface for pruning neurons, helps users understand the role of neurons through the filter visualization technique~\cite{springenberg2014striving}.
However, it describes its pruning approach mainly based on metrics that measure instability and sensitivity of neurons, ignoring groupings of semantically similar neurons.
We draw inspiration from the above important prior research to automatically find groups of similar neurons and interpret semantic similarities among them.

\subsection{Connection among Neurons}
A key role in neural networks is played by neurons.
Neurons are responsible for receiving and transmitting activation signals.
However, they do not work in isolation. 
For a neuron to detect a feature, it requires an orchestrated interaction among many neurons across different layers.
Recent research~\cite{olah2020zoom, hohman2019s, das2020bluff, das2020massif} visually explains how higher level concepts can be constructed by neural connections.
In the context of adversarial attack, some methods \cite{liu2018analyzing, das2020bluff, das2020massif} identify \textit{where} in a network the activation pathways of a benign and attacked input instance diverge, and how those diverging activations arrive at an incorrect prediction through connections among neurons.
Inspired by these techniques, we summarize and visualize how neuron groups interact through connections among them, providing a new way to interpret concept cascading across layers.

\section{Design Challenges}
\label{sec:challenge}
Our goal is to build an interactive visual summarization of concepts learned by neural networks. Concretely, we aim to help users better understand what concepts are represented internally by groups of neurons, and how these concepts are transformed into the final prediction through interactions among neuron groups. 
We identify the following five design challenges (\ref{challenge:similarity}-\remove{\ref{challenge:access}}\add{\ref{challenge:cascade}}) associated with developing our summarization techniques and designing \model{}.

\begin{enumerate}[label=\textbf{C\arabic*},itemsep=0mm, topsep=1mm,parsep=1mm, leftmargin=6mm]
    
    \item \label{challenge:similarity} 
    \textbf{Discovering neurons that detect similar concepts.}
    Existing research on DNN interpretability tends to focus on inspecting individual neurons~\cite{hohman2019s, olah2017feature, nguyen2016synthesizing}.
    While helpful, neuron-level inspection cannot easily reveal how clusters of neurons may detect the same concept, even though it is common for multiple neurons to detect similar features~\cite{he2017channel, jaderberg2014speeding, wen2016learning, duggal2019cup}.
    As a result, users can easily miss higher-order interactions that explain how DNNs operate.
    
    \item \label{challenge:relatedness}
    \textbf{Understanding the associations between related concepts.}
    
    Interpreting individual features represented by a neural network can be useful to understand what the model sees from input data~\cite{olah2017feature, nguyen2016synthesizing}.
    However, a model doesn't base its prediction on a single feature from the input. Instead, the final prediction is often an amalgamation of multiple concepts detected by the model. This raises fundamental questions about the associations between related concepts. 
    For example, when a concept is detected by a neural network (e.g., ``dog face''), what other concepts are likely to be detected at the same time, and how are they related (e.g., would ``dog tail'' and ``dog leg'' be strongly related to ``dog face'')? 
    
    \item \label{challenge:scalability} 
    \textbf{Scaling up concept summarization to all classes, neurons, and large datasets.} 
   
    Recent research in our visualization community has started to 
    prioritize scalability to support large datasets~\cite{hohman2019s, carter2019activation}.
    However, understanding how those approaches may extend to enable the discovery of groups of similar neurons---and encode semantic relatedness of concepts detected by the neurons---remains unclear.
    In the context of our work, computing neuron relationships can be computationally expensive.
    A naive algorithm to measure the neuron similarity would require comparing all neuron pairs, which is computationally expensive (i.e., time complexity is quadratic in the number of neurons).
    This naturally leads us to the question: how can we more efficiently support grouping neurons and encoding concept relatedness for complex DNNs?
    
    \item \label{challenge:cascade} 
    \textbf{Understanding concept influence in a network.}
    A promising approach to interpret a model's internal behaviour 
    involves understanding how 
    the model detects and combines features during inference \cite{olah2020zoom,liu2018analyzing}. 
    Recent research has proposed approaches to help users interpret how features may be connected~\cite{hohman2019s, das2020bluff}, but these approaches are performed at the neuron level, and are limited to only analyzing the relationships of neurons across two adjacent layers, instead of across the whole network.
    To this end, our work aims to answer a broader set of questions: How can a group of neurons detecting a concept trigger other concepts across the connections and layers of a DNN?
    Furthermore, how can we design an interactive visualization to support flexible exploration of such a ``concept cascade''?
    
\end{enumerate}

\section{Design Goals}
\label{sec:goal}

We distill the design challenges identified in \autoref{sec:challenge} to the following design goals (\ref{goal:similarity}-\remove{\textbf{G5}}\add{\ref{goal:association}})
that guide \model{}'s development.

\begin{enumerate}[label=\textbf{G\arabic*},itemsep=0mm, topsep=1mm,parsep=1mm, leftmargin=6mm]

    \item \label{goal:similarity} 
    \textbf{Clustering similar neurons based on activation overlap.}
    
    We aim to address a major research gap in existing work by developing techniques to discover neurons that detect the same features (\ref{challenge:similarity}).
    Specifically, we build on prior research findings that neurons tend to selectively respond to certain input features; 
    in the context of DNNs,
    this means such neurons' \textit{activation maps} have larger values at locations where the feature is present in the input image~\cite{long2014convnets, yosinski2015understanding}.
    Our idea is to group neurons based on how similar their activation maps are by (1) comparing the locations of the highly-activated values in the maps 
    (\autoref{sec:clustering})  
    and (2) visualizing the concepts that are detected by such neuron groups (\autoref{sec:graph}).
    
    \item \label{goal:relatedneess}
    \textbf{Encoding concept associations between related concepts.}
    We aim to analyze and visualize how concepts are related based on how often they co-occur (\ref{challenge:relatedness}).
    Our intuition is that neurons detecting highly related concepts (e.g., ``dog face'', ``dog tail'') are frequently co-activated.
    We aim to preserve these concept associations by learning vector representations for neurons that detect concepts associations on large image datasets (\autoref{sec:embedding}).
    Furthermore, we visualize the concept embedding to enable users to interactively explore and understand related concepts \add{across different level of abstraction such as ``dog face'' (higher level) and ``dog eyes'' (lower level)} (\autoref{sec:embeddingview}).
    
    \item \label{goal:scalability} 
    \textbf{Scalable summarization of concepts learned by a neural network.} 
    We aim to scale up neuron clustering (\ref{goal:similarity}) and neuron embedding (\ref{goal:relatedneess}) techniques to all neurons and all images by avoiding an explicit comparison of all neuron pairs (\autoref{sec:method}). 
    For scalable neuron clustering, we aim to project neurons' activation patterns into a reduced dimension and hash neurons into buckets by using these reduced projections as the key. Using this technique, we can hash similar neurons in the same bucket with high probability; more importantly, we can do this in time \textbf{linear} to the number of neurons instead of quadratic (\autoref{sec:clustering}).
    For scalable neuron embeddings, we aim to subsample neuron pairs and use the sampled pairs to learn neuron vectors that will preserve the general properties of concept relatedness. From these vectors, we can infer concept relatedness of any neuron pair without comparing them directly (\autoref{sec:embedding}).

    \item \label{goal:association} 
    \textbf{Interactive interface to explore \cascade{}}
    We aim to design and develop an interactive interface that
    enables users to selectively initialize and examine
    how a concept detected by a neuron group would trigger higher-level concepts across subsequent layers in a neural network \ref{challenge:relatedness}.
    \model{} visualizes the user-selected concept's cascade effect, helping users interpret the successive concept initiations and relationships (\autoref{sec:cascade}).
    
\end{enumerate}
\section{Model Choice and Background}
\label{sec:model}

In this work, we demonstrate our approach using \inception{}~\cite{mordvintsev2015inceptionism}, a prevalent, large-scale image classifier that achieves top-5 accuracy of 89.5\% on the ImageNet dataset that contains over 1.2 millions images across 1000 classes. 
\inception{} consists of multiple inception modules of parallel convolutional layers.
In each module, there are four layers: an input layer, an intermediate layer where kernels' size are 3x3, another intermediate layer where kernels' size are 5x5, and an output layer.
Each inception module is given a name of the form ``mixed\{number\}\{letter\}," 
where the \{number\} and \{letter\} denote the location of a layer in the network.
In \inception{}, there are 9 such modules: mixed3\{a,b\}, mixed4\{a,b,c,d,e\}, and mixed5\{a,b\}.
The input and output layer are given by the module name. 
For the intermediate layers, an suffix of either 3x3 or 5x5 is appended.
For example, mixed3b is an earlier input layer and mixed3b\_3x3 is an intermediate layer.
While there are more technical complexities 
in each inception module, we follow existing interpretability literature and consider the 9 mixed layers as the primary layers of the network~\cite{olah2017feature, olah2018building}. 
Although this work uses specific architectural choice, the proposed summarization and visualization techniques are general and can be applied to other neural network architectures in other domains.

\section{Scalable Neural Network Summarization}
\label{sec:method}

\model{} introduces two new scalable summarization techniques:
(1) \textit{neuron clustering} groups neurons based on the semantic similarity of the concepts detected by those neurons, and
(2) \textit{neuron embedding} encodes the associations between related concepts based on how often they co-occur.
\model{} leverages these techniques to summarize concepts learned by neural networks.
We formulate neuron clusterings in \autoref{sec:clustering}, describe neuron embeddings in \autoref{sec:embedding}, and detail how we filter concepts that are important for the prediction of each class in \autoref{sec:filter}.

\subsection{Neuron Clustering}
\label{sec:clustering}

We aim to discover groups of neurons within the same layer that detect the same concepts.
Our main idea is to group neurons that have similar activation maps. 
A neuron's activation map is a 2D image representing the  neuron's response to an input instance,
computed by the convolution of a trained kernel applied to the previous layer.
A neuron's activation map reflects features detected by a neuron, showing increased values in regions of the map where detected features exist.
Thus, if two neurons have similar activation maps, where highly activated areas of two activation maps largely overlap, we group the neurons.
For example, in \autoref{fig:neuron-selectivity}, two neurons 460 and 483 in layer mixed4c of \inception{}
are grouped by our approach, since their activation maps have high values on similar areas.
Our neuron clustering approach has two phases.
First, in the\remove{ prepossessing} \add{preprocessing} stage, we cluster neurons quickly and efficiently without looking at neurons' activation maps in detail.
Next, in the main clustering stage, we further divide the\remove{ prepossessed} \add{preprocessed} neuron groups based on the degree of overlap in the neurons' activation maps.

\subsubsection{Preprocessing: Group Neurons Based on Common Preferred Images}
\label{sec:preprocessing}

The preprocessing stage aims to efficiently and quickly cluster neurons before comparing neurons' activation maps in detail.
Our main idea\remove{ of the preprocessing stage} is to group neurons if they are highly activated by many common images.
For each neuron $i$, we first find a set of $k$ images that maximally activate $i$.
We sort the images by the maximum value of activation maps of $i$ for given those images, and take the first $k$ images.
We denote the set of top $k$ images for\remove{ neuron} $i$ as $X_i$.
For two neurons $i$ and $j$, we define their similarity as the Jaccard similarity of $X_i$ and $X_j$\remove{. The formal definition is} as follows.

\begin{definition} 
    \label{def:JS} 
    \textbf{Concept Similarity Based on Top Images.}
    \normalfont
    Given two neurons $i$ and $j$, and the neurons' top image sets $X_i$ and $X_j$, we define the similarity of $i$ and $j$ as the Jaccard similarity between $X_i$ and $X_j$. 
    This value is 0 when the two image sets are disjoint, and 1 when they are equal. 
    Neurons $i$ and $j$ are more similar when the Jaccard similarity is closer to 1.
    We formally define the similarity of $i$ and $j$ in Eq. \eqref{eq:presim}
    \vspace{-0.5em}
    \begin{equation}
        \label{eq:presim}
        \presim{}(i, j) = 
        \frac{
            \add{|}X_i \cap X_j\add{|}
        }{
            \add{|}X_i \cup X_j\add{|}
        }
    \end{equation}
\end{definition}
\vspace{-0.5em}

To scalably group neurons based on the common image sets, \model{} uses two techniques:
(1) \textit{Min-Hashing} efficiently approximates the Jaccard similarity between two neurons' top image sets; 
(2) \textit{Locality-Sensitive Hashing (LSH)} efficiently hashes similar neurons in terms of the Jaccard similarity into the same buckets with high probability.
\add{It is a popular technique to use Min-Hashing and LSH to efficiently estimate Jaccard similarity between two sets and find sets of similar items, due to their scalability and theoretical guarantees on the accuracy of finding nearest neighbors~\cite{broder1997resemblance, chum2008near, das2007google, gionis1999similarity, tamersoy2014guilt}.}

\textit{Min-Hashing.}
It is computationally costly to measure the Jaccard similarity between large sets due to the expensive set intersection and union operations that Eq. \eqref{eq:presim} involves.
Min-Hashing~\cite{broder1997resemblance} efficiently estimates the Jaccard similarity.
Let $h$ be a hash function that randomly maps the entire\remove{ images $\{\mathbf{x}_1, ..., \mathbf{x}_N\}$ to $\{\mathbf{x}_1, ... , \mathbf{x}_N\}$} 
\add{items $\{1, ..., N\}$ to $\{1, ... , N\}$} in one-to-one correspondence.
Let $h_{\text{min}}$ be a min-hash function that outputs the minimum value retrieved from the function $h$: 
for a set $S$, $h_{\text{min}}(S) = \min_{s \in S}(h(s))$. 
The key property of Min-Hashing is that the probability of the $h_{\text{min}}$ values of two sets being equal is equal to the Jaccard similarity between the sets. 
Formally, \add{for two sets $S_1$ and $S_2$, }\remove{ $Pr[h_{\text{min}}(X_i) = h_{\text{min}}(X_j)] = \presim(i, j)$} \add{$Pr[h_{\text{min}}(S_1) = h_{\text{min}}(S_2)] =$ Jaccard Similarity between $S_1$ and $S_2$}.
\add{For each neuron $i$, we define $I_i$ as the index of images in $X_i$. By using the theoretical property of Min-Hashing, $Pr[h_{\text{min}}(I_i) = h_{\text{min}}(I_j)] = \presim(i, j)$.}

\textit{Locality-Sensitive Hashing (LSH).}
Min-Hashing efficiently estimates the similarity of two neurons' top common images.
However, it is still computationally expensive to measure the similarity of all neuron pairs.
LSH is a scalable technique that finds reasonable approximations for grouping similar items without comparing all item pairs~\cite{gionis1999similarity, hu2013mutantx, rajaraman2011mining}.
For each neuron $i$ and its top image\add{s' index} set $\remove{X}\add{I}_i$, we produce $n$ Min-hash values $h_1(\remove{X}\add{I}_i), ..., h_n(\remove{X}\add{I}_i)$ with $n$ hash functions $h_1, ..., h_n$.
Then we partition the $n$ values into $b$ bands, each consisting of $r$ values, such that $n = b \times r$.
For each band, we hash neurons into the same buckets where $r$ hash values of such neurons are identical.
Then we finally cluster $i$ and $j$ in the same group, if they appear in the same bucket in at least one band.
\add{Theoretically, the probability that neuron $i$ and $j$ will hash to the same bucket in at least one of the $b$ bands is $1-(1-s^r)^b$, where $s$ is the true Jaccard Similarity between $I_i$ and $I_j$~\cite{rajaraman2011mining}.}

\subsubsection{Main Clustering: Group Neurons Based on Overlap of Activation Maps}
\label{sec:mainclustering}

While the preprocessing stage offers an efficient approach for preliminary neuron grouping, 
the main clustering stage performs finer clustering based on overlap of activation maps.
In the preprocessing stage\add{,} for example, neurons for ``cars'' and neurons for ``roads'' might be grouped together, as those concepts may frequently co-occur in the same images.
The main clustering stage further divides these neurons into different groups based on the concepts encoded in the activation map of the neurons.
Within a\remove{ prepossessed} \add{preprocessed} group, we finally cluster neurons in the same group, if highly activated part of the neurons' activation maps overlap significantly.
We formally define the similarity of neurons $i$ and $j$ used in the main clustering stage as follows.

\begin{definition} 
    \label{def:IoU} 
    \textbf{Concept Similarity Based on Activation Map.}
    \normalfont
    Given an input image $\mathbf{x}$ and two neurons $i$, $j$ in the same layer, 
    we denote their activation map as $Z_i(\mathbf{x})$ and $Z_j(\mathbf{x})$.
    To take only highly activated parts in each activation map, we quantize the activation maps as $Q_i(\mathbf{x}) = Z_i(\mathbf{x}) > 0$ and $Q_j(\mathbf{x}) = Z_j(\mathbf{x}) > 0$, where the quantized activation maps are a boolean matrix (i.e., true means high activation).
    We define the concept similarity of $i$ and $j$ in Eq. \eqref{eq:sim}, where $\land$ and $\lor$ are element-wise \texttt{and} and \texttt{or} operation respectively, and \texttt{numTrue}($\cdot$) returns the number of true values in the input matrix.
    If \add{\texttt{numTrue}(}$Q_i(\mathbf{x}) \lor Q_j(\mathbf{x})\add{)} = 0$, the similarity between $i$ and $j$ is defined as 0.
    \vspace{-0.5em}
    \begin{equation}
        \label{eq:sim}
        \actsim{}(i, j) = 
        \frac{
            \texttt{numTrue}(
                Q_i(\mathbf{x}) 
                \land 
                Q_j(\mathbf{x})
            )
        }{
            \texttt{numTrue}(
                Q_i(\mathbf{x}) 
                \lor 
                Q_j(\mathbf{x})
            )
        }
    \end{equation}
\end{definition}
\vspace{-0.5em}

The similarity $\actsim{}(i, j)$ in Eq.~\eqref{eq:sim} can be interpreted as the Jaccard similarity of highly activated parts in activation maps of $i$ and $j$.
We leverage Min-Hashing and LSH in the main clustering phase again for improved scalability.
For each \add{neuron} group $G$ created in the\remove{ prepossessing} \add{preprocessing} stage, we sample $t$ images from the union of every belonging neuron's top $k$ images.
We denote the set of such sampled images as $\remove{I}\add{X}_G$.
Formally, $\remove{I}\add{X}_G = \texttt{sample}(\cup_{i \in G} \remove{I}\add{X}_i, t)$, where $\texttt{sample}(S, t)$ randomly samples $t$ items in set $S$.
The main reason we use the sampled images (instead of all images) is that using all images is not very useful;
because neurons selectively respond to only some images, the neurons are not activated at all by many images.
To compare the similarity of two neurons, we only consider cases where both neurons are highly activated.
Thus, we sample images from the union of top $k$ images produced in the preprocessing stage, which includes many images that are likely to activate many neurons in the group $G$.
For each group $G$ produced in the preprocessing phase and for each image $\mathbf{x} \in I_G$, we run LSH to further group neurons based on the activation map. 
Then we finally group two neurons in the same bucket if the two neurons are hashed in the same bucket for least one image in $I_G$.
\remove{We empirically choose the following hyperparameters for our demonstration: 
$k=200$ images from 1.2M images,
$b=2000$, $r=3$ in the preprocessing stage,
and $b=5$, $r=20$ in the main clustering stage.}

\subsubsection{\add{Hyperparameter Selections for Neuron Clustering}}
\add{
Our neuron clustering approach uses a few hyperparameters: 
$t$ is the maximum number of sampled images for each preprocessed neuron group,
$k$ is the number of top images (among 1.2M images) for each neuron,
$b$ is the number of bands in LSH, and
$r$ is the size of the bands.
$t$ helps reduce runtime through sampling; a larger value means using more samples (thus longer runtime). We experimented with values in $[50, 200]$ and observed little change in the results, thus we decided on $t=100$. 
A larger $k$ 
increases the chances of discovering more neuron pairs that are similarly activated. However, a value that is too large (e.g., 1M) means most neurons would have highly similar or identical sets of top images.
We decided on $k=200$, the highest value that provided good clustering while keeping total runtime reasonable.
A larger $b$ provides more opportunities to group neurons that have high Jaccard similarities. 
For preprocessing,
we experimented with values in 
$[5, 2500]$ and the clustering results stabilized after $b$ reached 1500, thus we used $b=2000$. 
For main clustering, we experimented with values in 
$[5, 32]$, and used $b=20$ as 
clustering results did not change beyond that.
A larger $r$ allows us to prune neuron pairs with low Jaccard similarities. However, a value that is too large could prune neuron pairs even if they have high Jaccard similarities. Thus, we aimed to pick a value that is not too large, or too small. We experimented with values in 
$[2, 5]$ for preprocessing, and $[2, 30]$ for main clustering, and found the ``middle'' values of 3 and 15, respectively, provided good coherence among examples image patches in the cluster results.
}

\subsection{Neuron Embedding}
\label{sec:embedding}
To encode associations between concepts detected by neurons, we learn neuron embeddings that preserve the relatedness of such concepts,
\remove{Our goal is to learn vector representations}
where neurons that detect more related concepts are located closer in the embedding space.
\add{Our embedding approach consists of two steps.
\begin{itemize}[itemsep=0mm, topsep=1mm,parsep=1mm, leftmargin=3mm]
    \item \textbf{Step 1.} Learn vector representations of all neurons to encode relatedness among neurons' concepts. (\autoref{sec:embstep1})
    \item \textbf{Step 2.} Reduce the dimensions of the learned vector representation to 2D for visualization (\autoref{sec:embstep2}), which we will describe in \autoref{sec:embeddingview}.
\end{itemize}
The decision to adopt a two-step approach
to first generate higher-dimensional vector representations for neurons (Step 1) was motivated by prior work \cite{yin2018dimensionality, grover2016node2vec, mikolov2013efficient}, where abstract concepts are better captured by higher-dimensional representations, which opens up the possibilities for supporting interpretation tasks at higher fidelity.
}

\subsubsection{\add{Step 1:} Encode Relatedness of Neurons' Concepts via Vector Representations}
\label{sec:embstep1}
The objective function $J$ to minimize of our embedding approach is defined in Eq. \eqref{eq:objective}, where 
$D$ is a set of sampled neuron pairs detecting highly related concepts, 
$V_i$ is the embedding vector of neuron $i$ to learn, 
and $\sigma(x)$ is the sigmoid function \add{(i.e., $\sigma(x) = 1 / (1 + e^{-x})$)} \remove{defined in Eq. \eqref{eq:sigmoid}}.

\begin{equation}
    \label{eq:objective}
    J = \displaystyle
    \sum_{(i, j) \in D} 
        -\log(\sigma(V_i \cdot V_j))
\end{equation}

\noindent
The objective function induces the embedding vectors $V_i$ and $V_j$ of neurons $i$ and $j$ detecting highly related concepts to yield high $\sigma(V_i \cdot V_j)$.
A large value of the dot product of two vectors indicates that the vectors are far from the origin in the same direction.
The sigmoid function controls the magnitude of dot product, so that those vectors do not move too far away from the origin.
Thus, the objective function induces vectors of highly related neurons to be located closely and moderately far away from the origin.
Our use of cross-entropy loss was motivated by prior work~\cite{mikolov2013efficient} where no predefined classes or labels are available, which is the case here (i.e., no concept labels for each neuron).

To sample neurons of highly related concepts, we find neurons that are frequently co-activated.
We reuse the top $k$ images for each neuron which are obtained at the preprocessing stage of neuron clustering (\autoref{sec:preprocessing}).
For each image, we first generate a list of neurons that have such image in the neuron's top $k$ images.
Then, we sample neuron pairs from the top neuron list.
We first randomly shuffle the neuron list, 
apply a sliding window of size 2 on the shuffled list, 
and sample pairs of neurons that are co-occurred in the sliding window.
A good property of using sampled neuron pairs instead of all pairs is that sampled pairs indirectly can imply the relation of all neuron pairs.
If two pairs $(a, b)$ and $(b, c)$ of highly related items are sampled, we can infer that $(a, c)$ is also highly related. 
Our embedding approach efficiently learns and preserves such indirect concept relatedness, as well as direct relatedness straightly from the sampled pairs.
Note that the number of sampled pairs is linear to the number of neurons, not quadratic.
This sampling approach results in the training data of size linear to the number of neurons, and the time complexity of our neuron embedding method is linear to the number of neurons.

To further speed up the optimization process, we use negative sampling approach: concretely, we find pairs of non-related neurons and induce their embedding vectors far apart.
For a given neuron, we find another non-related neuron by randomly sampling one among all neurons in the same layer.
The new objective is defined in Eq.~\eqref{eq:updatedJ}, where $M$ is the size of negative sampling for a pair of related neurons $(i, j)$.

\vspace{-1.5em}
\begin{equation}
\begin{split}
    \label{eq:updatedJ}
    J = \displaystyle
    \sum_{(i, j) \in D} &\bigg(
        -\log(\sigma(V_i \cdot V_j)) \\
            &+ \sum_{m=1}^{M}
            \big(
                -\log(1-\sigma(V_i \cdot V_m))
                -\log(1-\sigma(V_j \cdot V_m))
            \big)
    \bigg)
\end{split}
\end{equation}
\vspace{-1em}

\noindent
We use gradient descent to learn neuron vector representations
that optimize $J$.
The derivatives of objective function $J$ with respect to the neuron vector $V_i$ and $V_j$ are as in Eq. \eqref{eq:deri1} and \eqref{eq:deri2}.

\vspace{-0.5em}
\begin{equation}
    \label{eq:deri1}
    \frac{\partial J}{\partial V_i} =
    -\big(1-\sigma(V_i \cdot V_j)\big)V_j
    + \sum_{m=1}^{M}\sigma(V_i \cdot V_m) V_m
\end{equation}

\begin{equation}
    \label{eq:deri2}
    \frac{\partial J}{\partial V_j} =
    -\big(1-\sigma(V_i \cdot V_j)\big)V_i
    + \sum_{m=1}^{M}\sigma(V_j \cdot V_m) V_m
\end{equation}

\noindent
We update the embedding by gradient descent as in Eq. \eqref{eq:update1}\remove{ and \eqref{eq:update2}}, where $\gamma$ is the learning rate. 
\remove{We use $\gamma$=0.01 with 30 epochs in our demonstration, since they lead to good quality of learned embeddings.}

\begin{equation}
    \label{eq:update1}
    V_i \leftarrow 
    V_i - \gamma \frac{\partial J}{\partial V_i}, 
    \;\;\;\;\;\;
    \add{
    V_j \leftarrow 
    V_j - \gamma \frac{\partial J}{\partial V_j}
    }
\end{equation}

\subsubsection{\add{Step 2: Dimensionality Reduction}}
\label{sec:embstep2}
\add{To project neurons' vector representations learned in the previous step onto a 2D space, we use UMAP, a non-linear dimensionality reduction technique that preserves global data structures and local neighbor relations~\cite{mcinnes2018umap}.}

\subsubsection{\add{Hyperparameter Selection for Neuron Embedding}}
\add{
Our neuron embedding uses a few hyperparameters:
size of negative sampling $M$,
number of epochs, and learning rate $\gamma$ for training.
Tuning $M$ helps prevent overfitting;
a value that is too large could introduce significant noise during training.
Epochs and learning rate affect the training runtime and quality.
More epochs usually causes more distinct clusters to form.}
We experimented with different combinations of hyperparameter values and found these chosen ones provide a good visual results: $M$=10, epoch=30, $\gamma$=0.01.

\subsection{Filtering Each Class's Important Model Substructures}
\label{sec:filter}

\subsubsection{Important Neurons and Neuron Groups}
\label{sec:node}

Our goal is to summarize important model substructures (i.e., neurons or neuron groups) that contribute to a model's class prediction.
To do so, we follow an approach similar to~\cite{hohman2019s}, and adapt our implementation to include neuron groups.
The importance of each neuron $i$ in layer $l$ for the prediction for a class $c$ is computed as the number of images of $c$ by which $i$ is maximally activated.
Whether a neuron $i$ in layer $l$ is highly activated for an image $\mathbf{x}$ is decided by the maximum value of activation map of $i$ for $\mathbf{x}$ (i.e., $\max(Z_i^l(\mathbf{x}))$).
For an image $\mathbf{x}$ for a class $c$, we find 5 most activated  neurons for each layer as suggested in~\cite{hohman2019s}. 

After obtaining the importance of each neuron for a class $c$, we then compute importance of each neuron group for $c$.
For a neuron group $G$, we take 10 neurons at most that have the highest importance for $c$, to avoid overcrowding the visualization, while we also observe that 10 neurons are enough to explain what concepts the group is detecting.
We then compute the importance of $G$ for $c$ as the average of importance score of at most the top 10 neurons.

\subsubsection{Important Connections among Neurons and Neuron Groups}
\label{sec:connection}

Important neurons and neuron groups summarize concepts that are important for a class prediction.
We further want to describe how those concepts interact to form higher level abstractions, by representing the connections between the model substructures.
We follow similar steps in \cite{hohman2019s} to compute the influence from each neuron in a given layer to each other neuron in a successive layer.
At a high-level, for a given class $c$, the influence of each neuron-neuron connection is the number of images of $c$ that use such connection as a major path to transmit high activation signal.
Thus, if two neurons have strong influence values, it means that the neuron in an earlier layer activated the other neuron in a subsequent layer for many images of the selected class.
Finally, to detect if an image uses a connection as a major path, we compute the maximum convolution value of activation map corresponding to the source neuron multiplied with a slice of learned kernel tensor between the two neurons.
We refer the reader to \cite{hohman2019s} for a more in-depth treatment of this approach. 
Our implementation deviates from \cite{hohman2019s} in the last step, when aggregating influence values for each neuron group.
For two neuron groups $G1$ and $G2$, we compute average of the influence values between any neuron in $G1$ and any neuron in $G2$, and use such average value as the connection weight between $G1$ and $G2$.

\section{User Interface}
\label{sec:interface}

Based on our design goals in~\autoref{sec:goal} and our neural network summarization techniques in~\autoref{sec:method}, we present \model{}, an interactive system that summarizes concepts detected by neuron clusters (\autoref{fig:teaser}).
The \model{} interface consists of three components:
(1) the header shows metadata and contains a few controls for the \graphview{},
(2) \embeddingview{} and \sideview{} provides a global overview of all neurons (\autoref{sec:embeddingview}), and
(3) \graphview{} visualizes concept relations (\autoref{sec:graph}).

\subsection{\embeddingview{} and \sideview{}: Global Overview of Neurons' Concept}
\label{sec:embeddingview}

The \embeddingview{} (\autoref{fig:teaser}A) aims to show a global overview of all neurons in a model, such that neurons detecting more related concepts are positioned closer.
We project embedding of all neurons computed in \autoref{sec:embedding} onto a 2D space.

\begin{figure}[t]
    \centering
    \includegraphics[width=0.85\columnwidth]{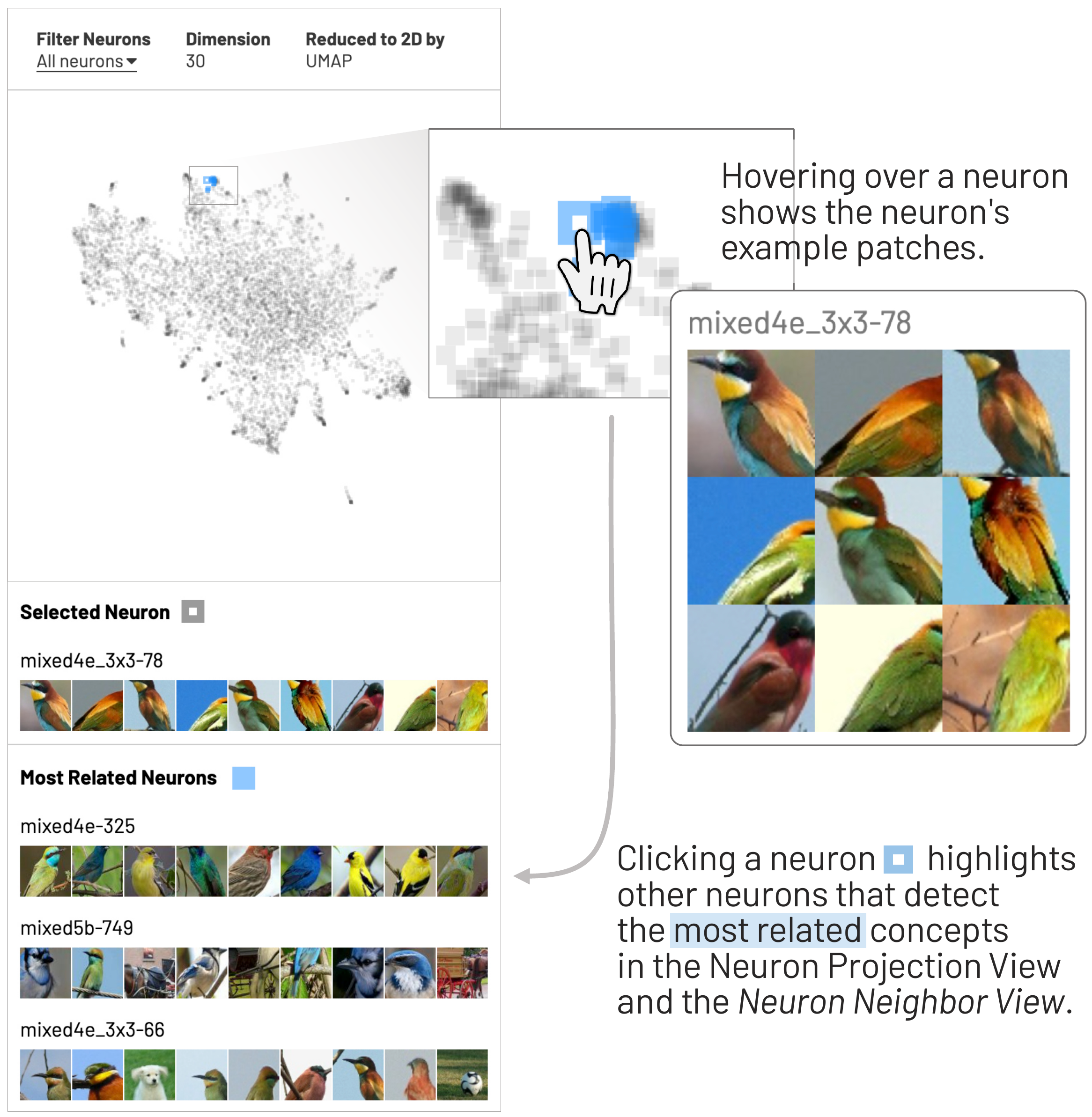}
    \vspace{-3mm}
    \caption{
        \embeddingview{} visualizes associations between related concepts based on co-occurrence, 
        by projecting neurons on a 2D space where each rectangle is a neuron.
        Hovering over a neuron shows its example data patches in a popup.
        Clicking a neuron selects it, marking it with a white dot at its center. 
        Related neurons are in blue, and related example patches are displayed in \sideview{} (at bottom left).
    }
    \vspace{-5mm}
    \label{fig:embedding-view}
\end{figure}

In the \embeddingview{}, each neuron is represented as a rectangle.
Hovering over a neuron shows representative example patches to explain such neuron's concepts (\autoref{fig:embedding-view}).
Users can focus on a neuron by clicking the corresponding rectangle. 
As a result, the selected neuron is marked with inner white rectangle, and the selected neuron's neighbors are highlighted with blue in the embedding space.
Also, at the bottom left, \sideview{} displays the neighbor neurons with their example patches (\autoref{fig:embedding-view}).
At the top, \embeddingview{} provides multiple filtering options, such as showing all neurons, neurons for a selected class, and neurons for selected clusters.
Users can freely zoom and pan in the view.

\subsection{\graphview{}: Neuron Clusters and their Interaction}
\label{sec:graph}

The goal of Graph View is to visualize what concepts are detected by which clusters of neurons, and how those clusters collaborate to form higher-level concepts and the final prediction (\autoref{fig:teaser}C).
\graphview{} provides two modes through toggle button in the header: 
(1) class exploration mode (\autoref{sec:class}) to visualize concepts important for a user-selected class
and 
(2) concept cascade mode (\autoref{sec:cascade}) to selectively activate a concept and examine its cascade effect.

\subsubsection{Class exploration}
\label{sec:class}
When a user selects a class in the header, \graphview{} shows a subgraph of the entire neural network that is relevant for the class prediction (\autoref{sec:filter}).
The nodes (circles) are neuron clusters or individual neurons within the same layers, displayed in the order of their important scores computed in \autoref{sec:node}.
The edges represent influence among the nodes, where edge weights are computed in \autoref{sec:connection}.
Thicker edges indicate more important connections. 
Users can filter the graph based on the importance score, using the a slider in the header.
The graph visualization is shown in a zoomable and panable canvas.

When a user hovers over a node, \model{} first highlights the node and the edges connected to the hovered node with pink, then displays the \clusterpopup{} (\autoref{fig:teaser}D) view, which contains example patches.
User can pin interesting nodes by clicking them.
The pinned nodes are highlighted in the \embeddingview{} and the \graphview{} with pink.
\embeddingview{} and \graphview{} are tightly integrated: hovering over a neuron in \embeddingview{} highlights its belonging cluster in the \graphview{}, and hovering over a node in \graphview{} highlights its member neurons in the \embeddingview{}.
Users can filter neurons in the \embeddingview{} to focus on pinned nodes using the dropdown menu in the header.

\subsubsection{\cascade{}: Successive Concept Detection Initiated by a User-Selected Concept}
\label{sec:cascade}

\begin{figure}[t]
    \centering
    \includegraphics[width=0.8\columnwidth]{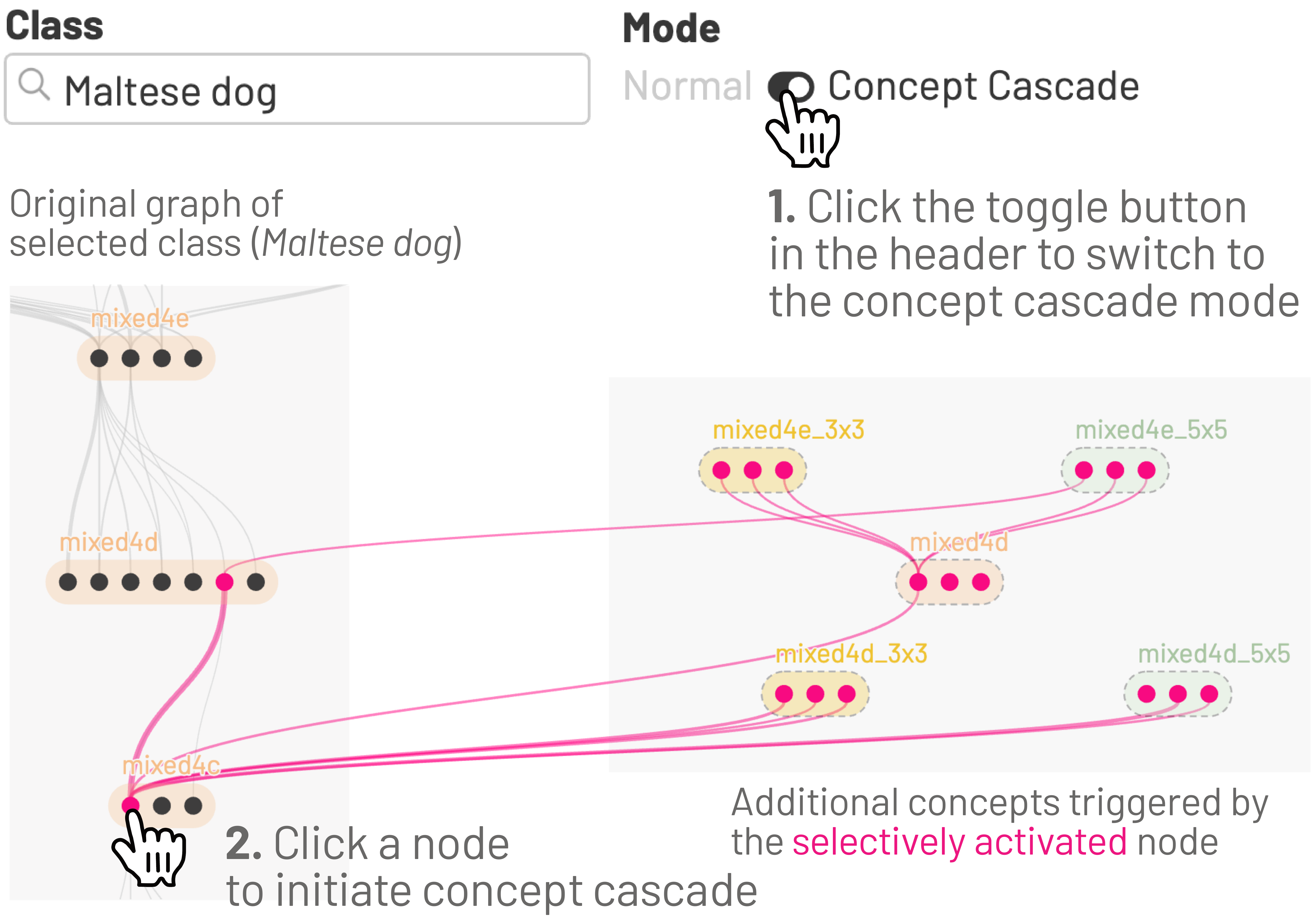}
    \vspace{-3mm}
    \caption{
Clicking a concept cluster node activates the concept, causing the neuron cluster to induce a \textbf{concept cascade} that triggers higher-level concepts across subsequent layers in the model.
\textbf{Left:} in the concept cascade, some neuron clusters are strongly contributing to current class's prediction and are shown in the class's graph summary. 
\textbf{Right:} Concepts less related are shown on the right hand side.
    }
    \vspace{-5mm}
    \label{fig:cascade-interface}
\end{figure}

Besides displaying how detected concepts interact within two adjacent layers for given images of a class, \graphview{} visualizes how one concept can influence multiple other concepts across all layers and classes through a concept cascade.
Users can enter the concept cascade mode by toggling the button in the header (\autoref{fig:cascade-interface}, at 1).
Then, users can click a concept cluster node to select it and manually activate the selected concept cluster (\autoref{fig:cascade-interface}, at 2), without feeding any input images. Clicking causes the neuron cluster to induce a concept cascade that triggers higher-level concepts across subsequent layers in the model.
Manual concept stimulation involves first setting every value in the activation map of the selected cluster's member neurons to 1, 
then forward-feeding such activation to the next layers through existing connections between neurons.
In the concept cascade, neuron clusters that are highly related to \textit{Maltese dog}, and strongly contribute to the prediction ``furry face,'' are included as part of the class's graph summary (\autoref{fig:cascade-interface}, left).
The cascade also includes concepts that related to the Maltese dog class, but are not as important for its prediction, such as
``bear face'' and ``black dog face.'' We highlight such concepts and their connections on the right side of the class's graph summary (\autoref{fig:cascade-interface}, right).

\subsection{System Implementation}
\label{sec:system}
To broaden access to our work, \model{} is web-based and can be
accessed from any modern web-browser. 
\model{} uses the standard HTML/CSS/JavaScript stack, and D3.js
for rendering SVGs. 
We ran all our deep learning code on a NVIDIA DGX 1, a workstation with 8 GPUs (each with 32GB of RAM), 80 CPU cores, and 504GB of RAM.
With this machine we could generate everything required for all 1000 ImageNet classes under 24 hours. 
The most computation intensive part was computing all neurons’ activation maps for all images (once for determining top-k images for each neuron; once for main clustering stage; each run was about 5 hours). Each of the other processes, thanks to the scalability of Min-Hashing, LSH, and the sampling-based neuron embedding, took less than an hour.

\section{Human Experiment to Evaluate \model{}}
\label{sec:study}

\begin{figure*}
    \centering
    \includegraphics[width=0.8\textwidth]{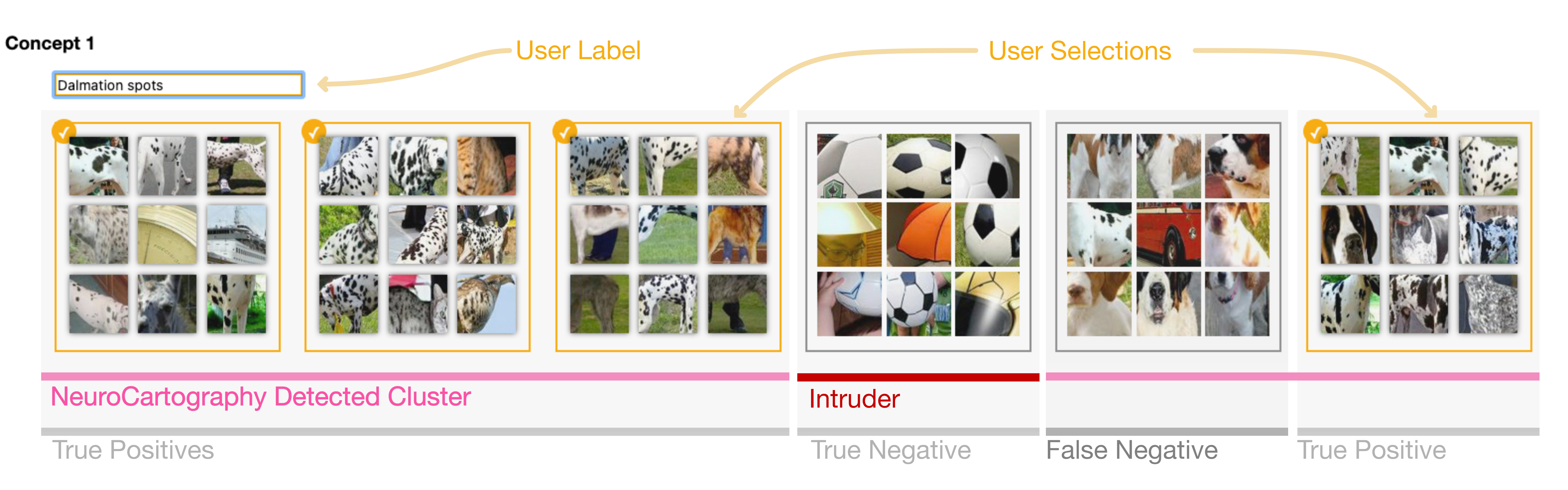}
    \vspace{-5mm}
    \caption{Example question from MTurk evaluation. Users were presented with six neuron patches and asked to determine if there is a coherent cluster and if so provide a short label. In this example of a \model{} generated cluster we can see which neuron is the out of cluster intruder, which neurons are in the cluster, which options the user selected, and the classification results of true positives for the neurons the user correctly selected, true negative for not selecting the intruder, and a false negative error for not selecting a neuron that is in the cluster.}
    \vspace{-5mm}
    \label{fig:mturk}
\end{figure*}

To validate the human interpretability of the clusters discovered with \model{}, we conducted a large-scale human evaluation using Amazon Mechanical Turk, a standard practice for computer vision tasks \cite{deng2009imagenet}, basing our experimental design on similar work in image \cite{ghorbani2019towards} and language \cite{chang2009interpret} based cluster interpretability studies. For the experiment each user was presented a series of tasks like the task shown in \autoref{fig:mturk}. In each task we display the example patches for 6 neurons which are composed of 6 randomly sampled neurons, or 5 neurons from a cluster (determined either by \model{} or hand selected) as well as one `intruder' neuron which is randomly sampled from the rest of the network. Users were told that each task contains either all random patches, or a cluster of 3-5 related neurons, and were asked to identify which, if any, of the shown patches form a cohesive cluster and could select any number of the given options. By not disclosing the number of intruders users are forced to only select clusters which are fully coherent among themselves (as opposed to simply finding the least representative example).  We can then measure `false positives' (where users mistakenly identify the intruder as part of the cluster) and `false negatives' (where users may decide to not include patches from the cluster). This style of study measures the performance of human annotators against model output as ground truth, the inverse of standard machine learning metrics, in order to understand the difficulty of the interpretation task that the interpretability method (in our case clusters from \model{}) provides. We assume that higher performance by humans equates to an easier task for humans to interpret, highlighting a better methodology for providing explainability.  Our evaluation specifically takes the inverse framing of other studies which ask users to positively identify the intruder rather than the cluster \cite{chang2009interpret, ghorbani2019towards}. We do this in order to independently measure the `false negative' class of errors that are not possible to detect using pure intruder detection. 

For our study, we generated 99 unique sets of neurons such as those in \autoref{fig:mturk}, of which 43\% were generated using \model{}, 43\% were generated from hand picked clusters, and 14\% were generated completely at random with no underlying cluster. These sets were used to populate 9 different questionnaires of 11 sets which Amazon Mechanical Turk workers located within the U.S. completed, receiving compensation of \$1  per questionnaire and taking an average time of 7 minutes to complete. An average of 42 unique workers completed each questionnaire, for a total of 3374 unique human judgements of clusters from 244 unique workers overall.  
\begin{figure}
    \centering
    \includegraphics[width=0.6\columnwidth]{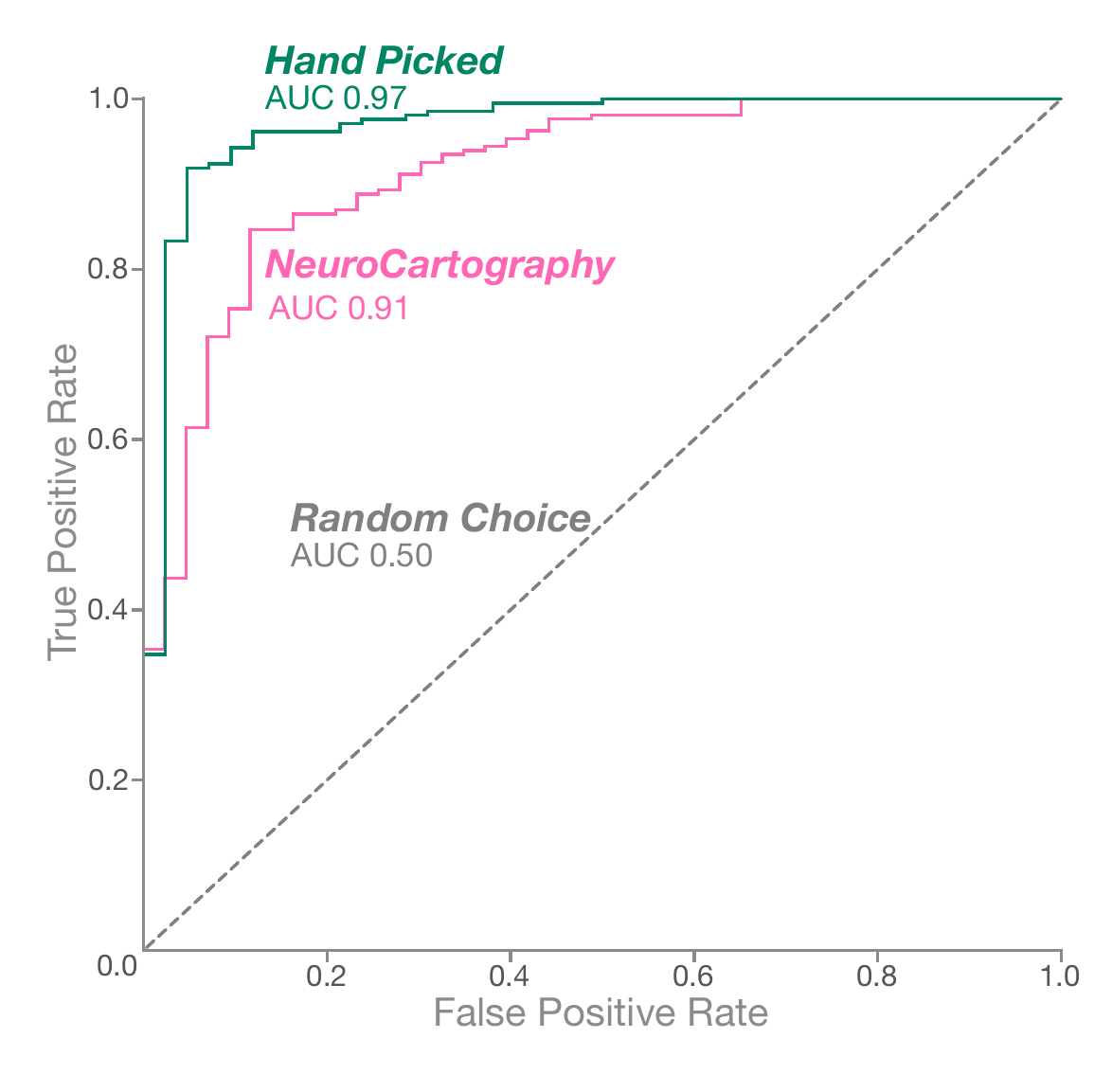}
    \vspace{-5mm}
    \caption{ROC Curve for user estimations of cluster inclusion. Both hand picked and \model{} generated clusters perform well overall, implying that the clusters generated are interpretable enough to be consistently recognised by different users.}
    \vspace{-5mm}
    \label{fig:roc}
\end{figure}

\subsection{Cluster Cohesion}
The measure of cluster cohesion used in other intruder detection experiments is accuracy on the binary prediction task on whether a user correctly identified the intruder. In our study this is equivalent to the false positive rate for cluster inclusion predictions. For the random baseline, the false positive rate was found to be $30.6\%\pm3.1\%$ (showing how easy it is for humans to find spurious patterns where none exist) while hand picked clusters had a false positive rate of $6.1\%\pm1.2\%$, and \model{} generated clusters had a false positive rate of $11.8\%\pm1.7\%$. With $95\%$ confidence intervals, both clustering techniques significantly outperformed the baseline on the binary task.  

However, in real applications, and in the context of this study, users have varying thresholds for how similar neurons must be in order to be included within the same cluster, even if they are using the same underlying similarity heuristic for their judgements. Because our study design gave participants a choice for the number of options to select within a given set, each neuron's inclusion is essentially an independent judgement of whether it fits within the cluster (if the user determines a cluster exists). By getting enough of these independent measurements for each neuron we can use the proportion of users who choose to include a neuron within a cluster as a score of how likely users expect it to be included, or how cohesive the specific neuron is with the the context of the whole cluster. By using this score instead of binary accuracy, we can evaluate the full space of trade-offs between false negative and false positive errors using the receiver operative characteristic (ROC) which provides a fuller, threshold-independent picture of performance, offering a more robust variation of the specific experimental setup and balance of options and intruders \cite{bradley1997roc}. 

The score used to define the ROC in \autoref{fig:roc} was calculated using the percentage of users including a given neuron within a cluster in the case that the user determines there is a cluster present in the set (that is they select 3, 4, or 5 neurons as opposed to 0). We used this method for both hand selected and \model{} generated clusters (and excluded the random baseline as it contains no true positive values). Taking the area under the curve (AUC) of the ROC for each clustering method (\autoref{fig:roc}) we find again that hand selected clusters again outperform \model{}, but that both perform substantially better than chance with AUC values of $0.97\pm 0.04$ for hand created clusters and $0.91\pm0.04$ for \model{}. This result shows that \model{} produces clusters that are nearly as interpretable as hand crafted clusters across different inclusion thresholds, and are both much more reliably detected than chance.

\subsection{Label Cohesion}
To further understand the consistency of the patterns agreed upon by users, we asked users to describe individual clusters they selected. 
Without a ground truth for cluster descriptions, we looked to statistically compare how different users labeled the same clusters in order to see the consistency of the discovered concept. To compare cluster level descriptions, we rely on sentence level embeddings from the Universal Sentence Encoder (USE)\cite{cer2018universal}. USE works by projecting sentences into embeddings which can be compared (using cosine similarity) to identify the presence of similar ideas. 
USE similarity is preferred to word choice overlap metrics used in \cite{ghorbani2019towards}, since it captures semantic similarity of the actual concepts being discussed regardless of phrasing (for example matching labels describing the same set with ``dots'', ``circles'', and ``round objects'').

First, in order to build a baseline for semantic similarity values, we calculated the average pairwise similarity between all labels across different clusters throughout the dataset to be $0.301\pm.003$. Then for each unique set, we found the average pairwise similarity between the labels from all of the users with labels for that specific set. These average within-group similarities for each class of cluster are $0.59\pm0.05$ for hand picked clusters and $0.51\pm0.05$ for \model{} generated clusters. Both of these results are significantly higher than the baseline, showing semantic consistency between how users understand the clusters that they detect. \add{Complementary future evaluation may assess the degree to which the neuron groups are capturing redundant semantics (e.g., track accuracy changes as neurons are pruned).}

\section{Usage Scenarios}
\label{sec:scenario}

\subsection{Automatically Discovering Backbone Concept Pathways for Related Classes}

DNNs are known to learn general concepts: this generalizability is widely leveraged in transfer learning, model compression, and model robustness research \cite{kawaguchi2017generalization, neyshabur2017exploring}.
However, it is challenging to \textit{automatically} discover and interpret which key concepts are progressively combined or connected internally in a model, or how such ``backbone'' concept pathways may be shared across related classes.
Recent research has proposed approaches
to help users interpret how 
features may be connected
~\cite{hohman2019s}, but
such approaches are performed at the neuron level, limited to only analyzing the relationships of neurons across two adjacent layers, instead of across the whole network. The biggest drawback is the dependence on manual processes.

\begin{figure}[t]
    \centering
    \includegraphics[width=0.7\columnwidth]{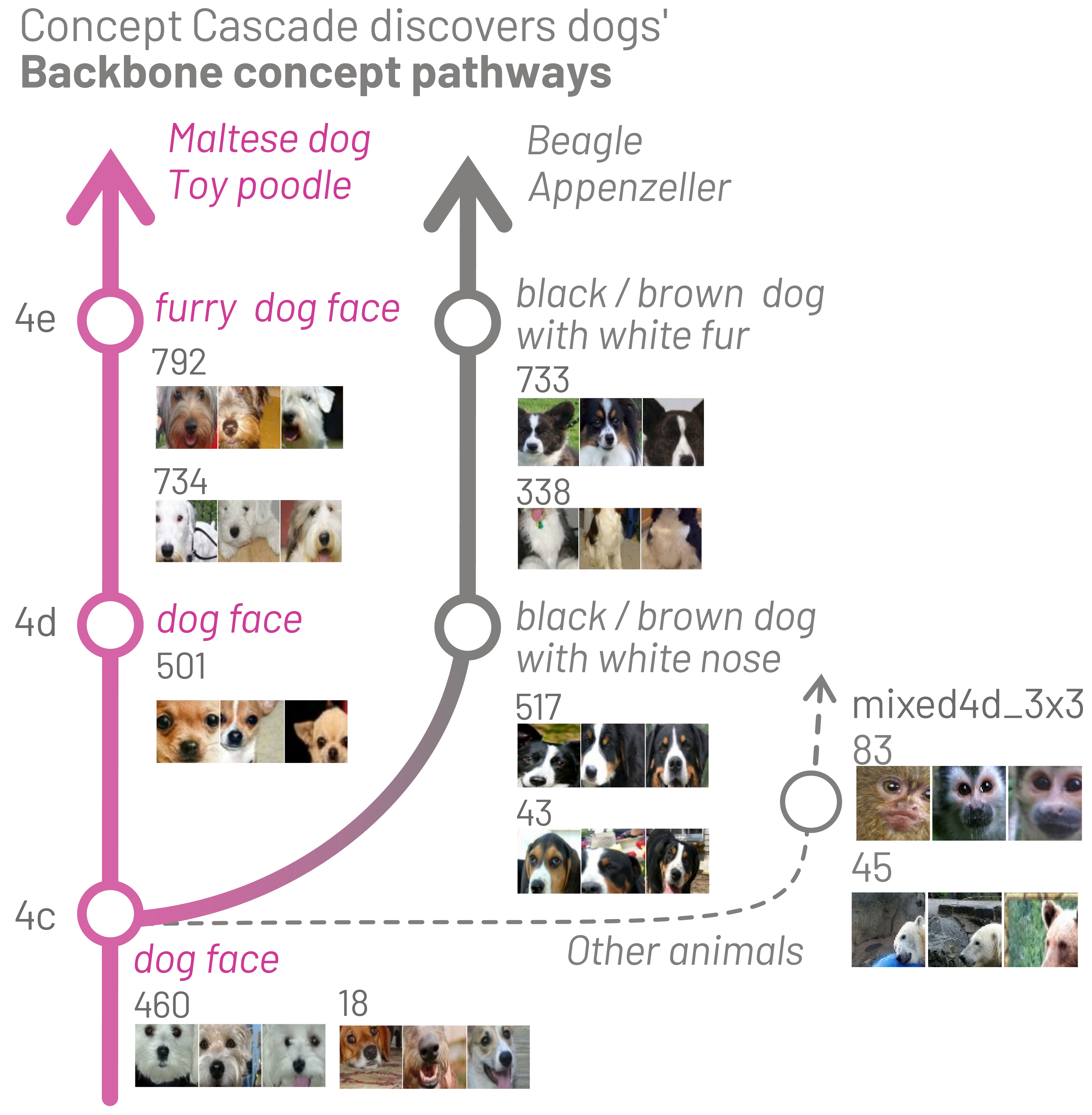}
    \vspace{-3mm}
    \caption{
        Through \cascade{}, users manually activate the ``dog face'' concept to discover its related concepts in subsequent layers, not only for the current ``Maltese dog'' class but also for other breeds of dogs. \cascade{} helps users visualize how concepts may evolve over the network, such as from the more generic ``dog face'' concept to the more specific ``furry dog face'' concept in later layers. Concepts are manually labeled.
    }
    \label{fig:cascade-ex1}
\end{figure}

\model{}'s \cascade{} can help automatically discover backbone concept pathways for related classes across all layers.
Users can selectively activate a concept and examine the concept's cascade effect to interpret the successive concept initiations in layer layers, while identifying concepts' general relationships to related classes.
For example, while inspecting the \textit{Maltese dog} class, we found a cluster detecting ``dog face'' in mixed4c (\autoref{fig:cascade-ex1}).
Through \cascade{} mode, we manually activate this concept to trigger and discover its related concepts in subsequent layers not only for the Maltese dog class, but also for other breeds of dogs such as \textit{Beagle} and \textit{Appenzeller}. 
Through these concept cascades, we visualize how concepts may evolve over the network, such as from the generic ``dog face'' concept to the specific ``furry dog face'' concept in later layers.

\begin{figure}[t]
    \centering
    \includegraphics[width=0.8\columnwidth]{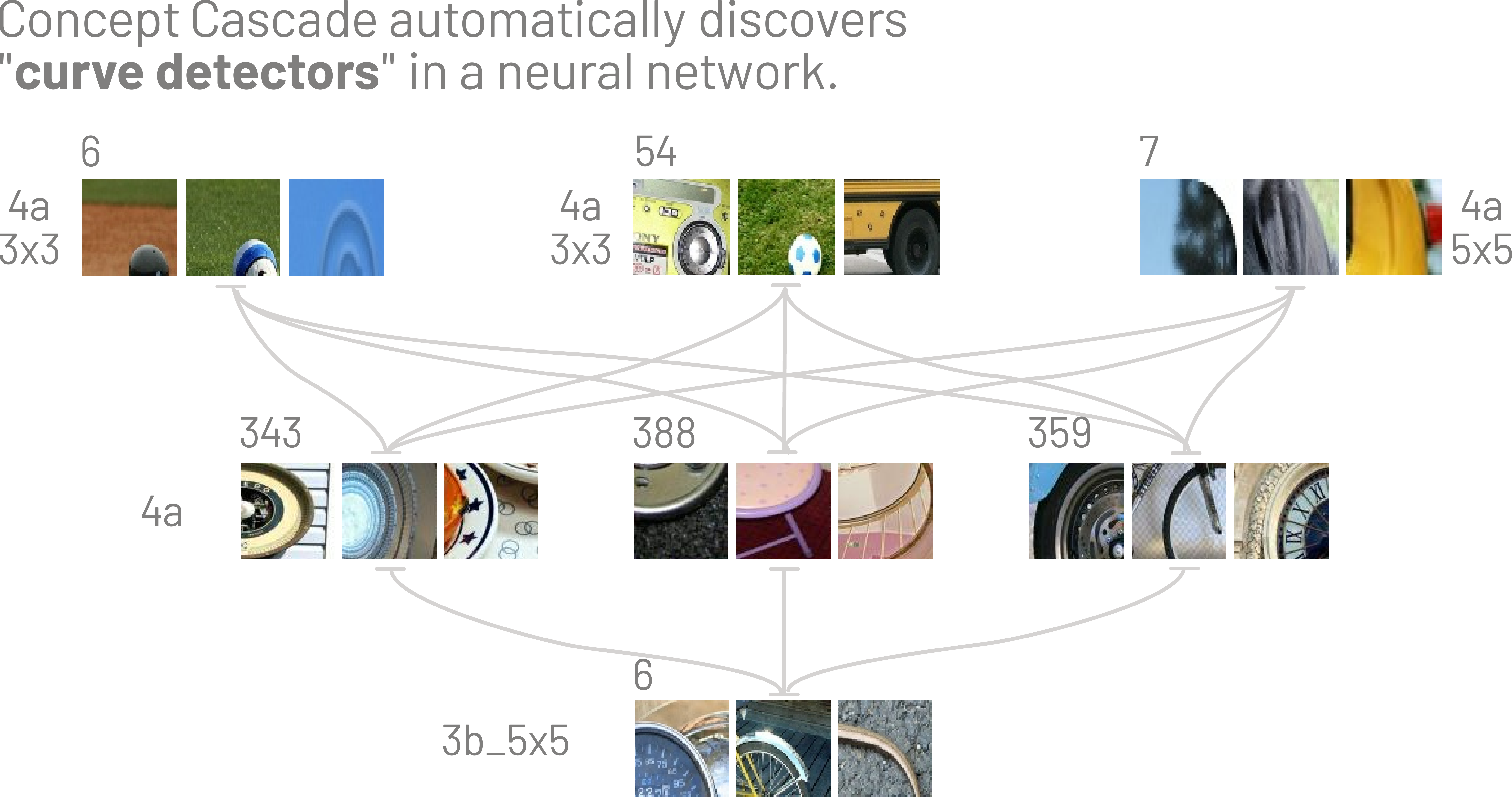}
    \vspace{-3mm}
    \caption{
        \cascade{} automatically discovers neurons that detects curve of specific orientations, which have been manually found in \cite{cammarata2021curve}.
    }
    \vspace{-5mm}
    \label{fig:cascade-ex2}
\end{figure}

Backbone concept pathways can also be used to highlight learned generalize properties, like ``curve detectors'' (\autoref{fig:cascade-ex2}).
Existing work \cite{cammarata2021curve} has observed that several neurons in the earlier layers detect curves of different orientations.
Even though these detectors were discovered manually,
their analysis yields interesting properties of the curve concepts, like how sensitive the curve detectors are to curvature and what orientations do they respond to.
Using \model{}, we automatically discovered more curve detectors in \inception{}.
We selected a node in mixed3b\_5x5 layer which detects a curve of a specific orientation, selectively activated in the \cascade{} mode, and discovered the curve detectors across layers such as neurons shown in \autoref{fig:cascade-ex2}.
We also observed that those curve detectors are clustered in the preprocessing stage of our clustering algorithm, but not grouped by the main clustering stage.
This is because the curve detectors are highly selective for orientations, causing highly activated regions of the activation maps are different by the detectors.

\subsection{Finding Isolated Concepts}
\begin{figure}[t]
    \centering
    \includegraphics[width=0.6\columnwidth]{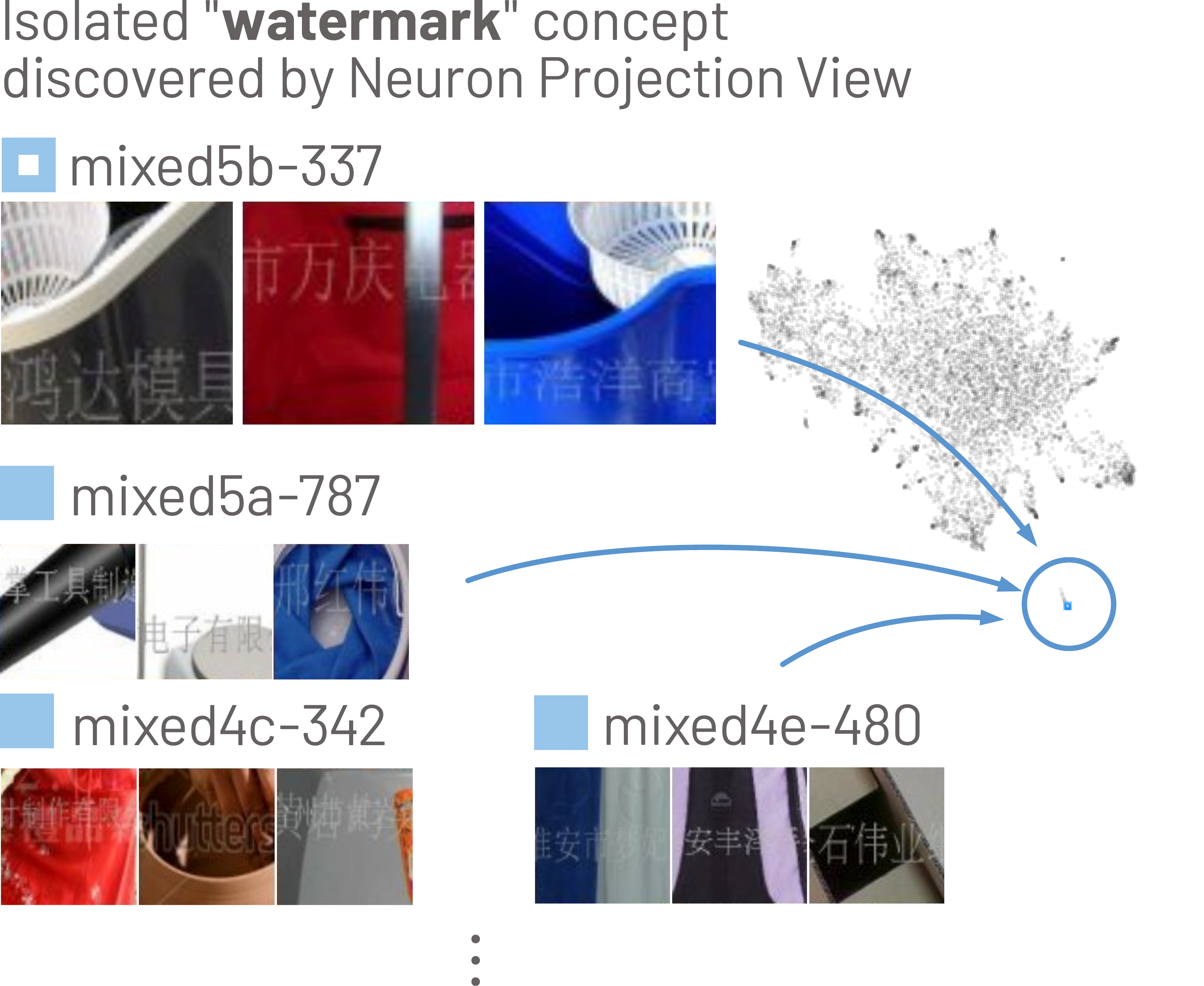}
    \vspace{-3mm}
    \caption{
        \model{} reveals the interesting phenomenon about the isolated ``watermark'' concept (example image at top-left), that
        watermarks are not specific to any image features (i.e., can appear on almost any kinds of images), 
        thus watermark neurons are placed far away from all other neurons due to relatively low co-activations (see \embeddingview{} on the right).
    }
    \vspace{-5mm}
    \label{fig:isolated}
\end{figure}

While inspecting the \embeddingview{}, we noticed a small number of neurons are distinctively positioned very far away from all other neurons in the embedding space (\autoref{fig:isolated}, right).
Inspecting such isolated neurons reveals that they are detecting the ``watermark'' concept (see example from mixed5b-337 layer at \autoref{fig:isolated}, top-left).
Interestingly, as watermarks can appear on almost any kinds of images independent of the image content that the watermarks are placed above (e.g., copyright watermark can appear on an image of a car, a dog, or a pineapple), 
this means the neurons responsible for detecting watermarks would frequently co-activate with each other, 
but such ``watermark neurons'' co-activate relatively less so with the neurons detecting the concepts that describe the image content since watermarks are not associated with only some specific features.
\model{}'s neuron embedding algorithm is able to discover this interesting phenomenon about the watermark neurons, placing them close together to reflect the concept coherence for watermark, 
and away from other neurons to reflect the watermark's non-specificity for image content.
\model{} allows us to easily verify our observations and conclusions.
For example, selecting mixed5b-337, a watermark neuron (\autoref{fig:isolated}, top-left), in the \embeddingview{} 
brings in its most related neurons in the \sideview{} (e.g.,mixed5b-86, mixed4c-342, mixed4e-296), which are all watermark neurons as well.
These neurons are also clustered in the \graphview{} 
(e.g., in mixed5b layer, neurons \#337, \#113, \#289, and \#86 appear in the same neuron cluster).

\section{Conclusion, Limitations and Future Work}

We have presented \model{}, an interactive system that scalably summarizes and visualizes concepts learned by DNNs via scalable concept summarization techniques for \textit{neuron clustering} and \textit{neuron embedding}.
Through a large-scale human evaluation, we have demonstrated that our techniques discover neuron groups that represent coherent, human-meaningful concepts.
Our system runs in modern browsers and is open-sourced.
Below, we discuss limitations of our approach and future research directions for extending this investigation.

\medskip
\noindent
\textbf{Further dissecting poly-semantic neurons.}
We believe our work has taken a major step in  addressing the research challenging of automatically and scalably grouping neurons that detect the same concept, going beyond manual, neuron-level inspection (e.g., \cite{bau2020understanding, hohman2019s, fong2018net2vec, nguyen2016multifaceted}) to provide a higher-level perspective for the knowledge learned by a network. 
Our work, however, is not designed for ``dissecting'' neurons that may become activated for multiple seemingly unrelated concepts, which has been observed in recent work, e.g, \cite{olah2020zoom}.
For example, in InceptionV1, at least poly-semantic neuron that responds to cat faces, fronts of cars, and cat legs \cite{olah2020zoom}.
\model{} cannot ``split'' this neuron into multiple neuron, each detecting one concept and put that newly created neuron into its logical neuron cluster with other similar neurons in the network. Tackling poly-semantic neurons is an exciting and challenging direction for future work.

\medskip
\noindent
\textbf{Integrating \model{} into more applications.}
Currently, our work focuses on using \model{} to enhance interpretability of DNNs. As DNNs are increasingly used in an ever-increasing variety of applications, our approaches can help practitioners and researchers assess the effectiveness of their ideas.
For example, in the neural network compression community, several methods~\cite{duggal2019cup, he2017channel, jaderberg2014speeding, wen2016learning} leverage potential neuron redundancies to generate compressed models while maintaining prediction accuracy.
\model{} can help researchers interpret the semantic similarity between the compressed model and the original, uncompressed models, which helps them assess if their techniques are indeed preserving the ``gist'' of the knowledge important for prediction, or if they are leveraging some other features of the data of the model.
\add{Currently, concepts need to be manually labeled; automatic labeling will increase the tool's usability.}
\add{Also, current \embeddingview{} presents all neurons in the same plot even though some concepts' abstraction levels could be very different; our future work includes providing users with the ability to select layers that they want to investigate.}
We look forward to seeing the impact that \model{} may contribute, from assisting evaluation of existing techniques (e.g., model compression, adversarial attacks and defenses), to developing new ones.

\medskip
\noindent
\textbf{Visualizing other neural network models.} 
We have justified our model choice in \autoref{sec:model};
we are working to extend support to other CNN models.
Our approach can easily be adapted to simpler models (e.g., VGG \cite{simonyan2014very}).
For more complex networks (e.g., ResNets \cite{he2016deep}, 
small extensions would be needed to handle more types of connections present in the network (e.g., skip connections could be represented as skip-layer edges in the graph view).

\acknowledgments{
We thank Hannah Kim, the Georgia Tech Visualization Lab, and the anonymous reviewers for their support and constructive feedback.
This work was supported in part by DARPA (HR00112030001), NSF grants IIS-1563816, CNS-1704701, and gifts from Intel, NVIDIA, Google.
}

\bibliographystyle{abbrv-doi}

\bibliography{999-ref}
\end{document}